\documentclass{article}

\PassOptionsToPackage{numbers, compress}{natbib}

\usepackage[preprint]{neurips_2022}

\usepackage[utf8]{inputenc} 
\usepackage[T1]{fontenc}    
\usepackage{hyperref}       
\usepackage{url}            
\usepackage{booktabs}       
\usepackage{amsfonts}       
\usepackage{nicefrac}       
\usepackage{microtype}      
\usepackage{xcolor}         

\usepackage{orcidlink}
\usepackage{graphicx}
\usepackage{amsmath}

\usepackage{graphicx}
\usepackage{subcaption}
\usepackage{multirow}

\usepackage{arydshln}

\usepackage{enumitem}

\newcommand{\tbf}[1]{\textbf{#1}}

\title{xGen-VideoSyn-1: High-fidelity Text-to-Video Synthesis with Compressed Representations}

\usepackage{authblk}
\author{Can Qin$^{*}$, Congying Xia$^{*}$, Krithika Ramakrishnan$^{*}$, Michael Ryoo$^{*}$, Lifu Tu$^{*}$, Yihao Feng$^{*}$, \\ 
Manli Shu, Honglu Zhou, Anas Awadalla, Jun Wang, Senthil Purushwalkam, Le Xue, 
Yingbo Zhou, \\ Huan Wang, Silvio Savarese, Juan Carlos Niebles$^{*}$, Zeyuan Chen$^{*}$, Ran Xu$^{*}$, Caiming Xiong$^{*}$ \vspace{2mm} \\ 
Salesforce AI Research \\
\noindent $^{*}$Core authors \\
\texttt{\{cqin, zeyuan.chen, ran.xu, cxiong\}@salesforce.com}
}
\affil{}

\begin{document}

\maketitle

\begin{abstract}
We present xGen-VideoSyn-1, a text-to-video (T2V) generation model capable of producing realistic scenes from textual descriptions. We extend the latent diffusion model (LDM) architecture by introducing a video variational autoencoder (VidVAE). Our Video VAE compresses video data spatially and temporally, significantly reducing the length of visual tokens and the computational demands associated with generating long-sequence videos. To further address the computational cost, we propose a divide-and-merge strategy that maintains temporal consistency across video segments. Our Diffusion Transformer (DiT) model incorporates spatial and temporal self-attention layers, enabling robust generalization across different time frames and aspect ratios.We also designed a data collection and processing pipeline, which helped us gather over 13 million high-quality video-text pairs. The pipeline includes steps such as clipping, text detection, motion estimation, aesthetics scoring, and dense captioning based on our xGen-MM video-language model. Training the Video VAE and DiT models required approximately 40 and 642 H100 days, respectively. Our model supports over 14-second 720p video generation in an end-to-end way and demonstrates competitive performance against state-of-the-art T2V models.
\end{abstract}

\section{Introduction} 
Text-to-video (T2V) generation models are designed to create videos that depict both realistic scenes from textual descriptions. These models are attracting attention both academia and industry due to recent breakthroughs.
Recently, Sora~\cite{videoworldsimulators2024} demonstrated that it is possible to generate realistic videos of over one minute in length. Despite such impressive advancements,
the most capable video generation models remain proprietary and their details undisclosed.
While many open video generation models have surfaced recently, their performance lags behind proprietary models.
Our goal with this work is to develop a highly effective architecture for text-to-video (T2V) that rivals existing state-of-the-art models. We examine associated modeling and training technologies, as well as explore the data-collection pipeline.

\begin{figure}[t]
\centering
    \includegraphics[width=\textwidth]{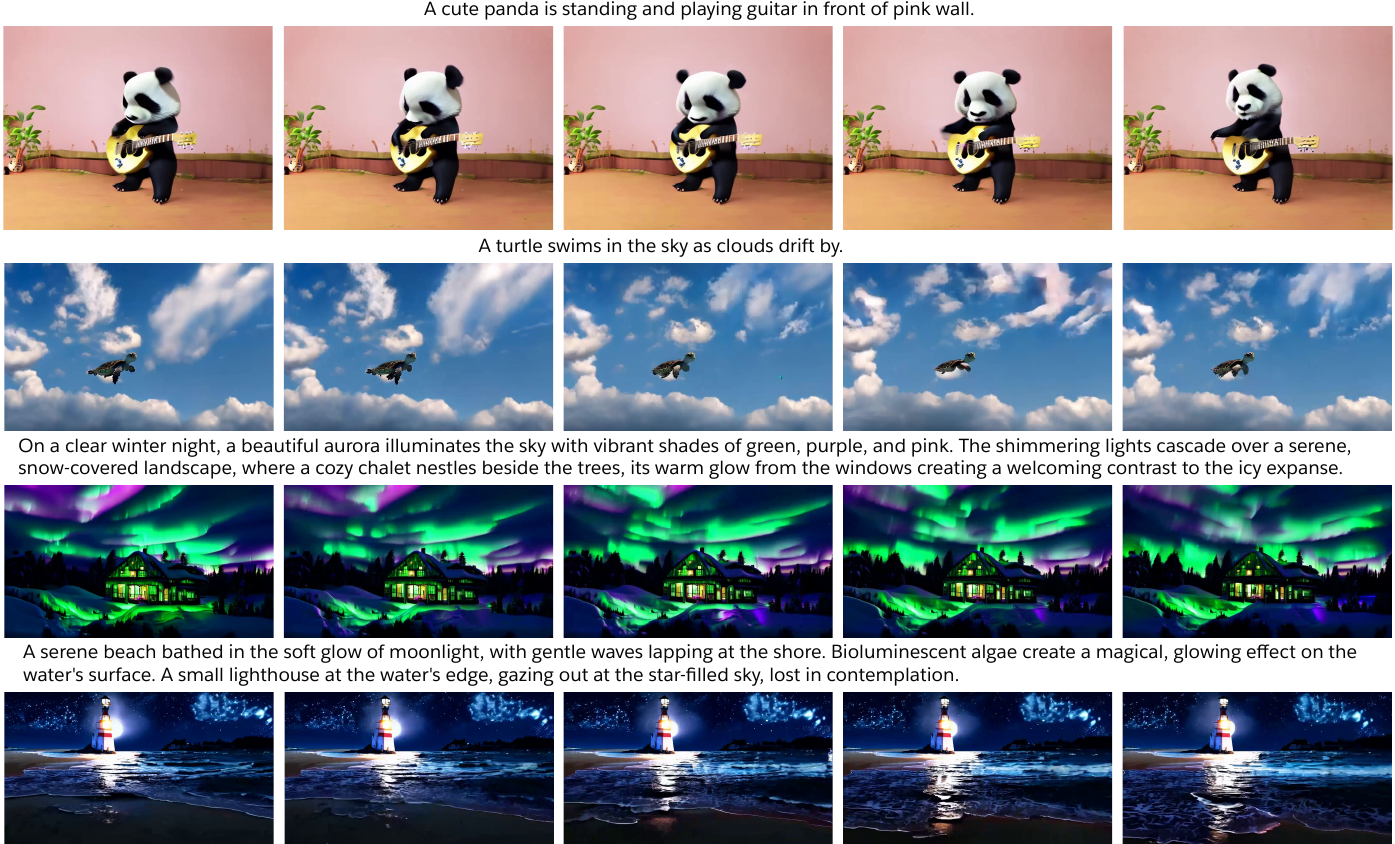}
    \caption{Example 720p text-to-video generation results by our xGen-VideoSyn-1 model.}\label{fig:result-1}
\end{figure}

A popular approach for image and video generation builds upon the latent diffusion model (LDM) \cite{rombach2022high} architecture. In this framework, pixel information is typically compressed with a pre-trained VAE \cite{kingma2013auto} into a latent encoded space. A diffusion process is then applied to this latent space either with a U-Net \cite{ronneberger2015u,ho2020denoising} or DiT architecture \cite{peebles2023scalable}. Generally, this framework has been adapted to both text to image~\cite{rombach2022high,sauer2024fast,chen2023pixart,pernias2023wuerstchen} and text to video~\cite{blattmann2023stable,ma2024latte,wang2023lavie} generation tasks.

A crucial component of such design is the dimensionality of the latent space determined by the output of the VAE. A latent space with small dimensionality means that the input pixel information needs to be highly compressed, which makes the reconstruction by diffussion more difficult but computationally less expensive. A latent space with large dimensionality makes reconstruction easier but computationally more expensive. In the case of image generation one can choose larger encoding spaces~\cite{rombach2022high} to facilitate reconstruction quality. However, this trade-off is particularly critical for video generation. If we encode each frame independently using an image VAE \cite{blattmann2023stable,ma2024latte}, a 100 frame video of 720p spatial resolution would translate into a latent space of size 100$\times$4$\times$90$\times$160 that contains $360000$ tokens. This makes both training computationally very expensive and inference slow.

To address this issue, we focus on developing a text-to-video (T2V) generation model based on video-specific VAE and DiT technologies.
We introduce a video VAE to achieve effective compression of the video pixel information by reducing both spatial and temporal dimensions. That is, instead of encoding each frame independently, we incorporate both temporal and spatial compression. This significantly decreases the token length, improves the computational cost of training and inference, and facilitates the generation of long videos. Additionally, to further reduce computation during long video encoding, we propose a divide-and-merge strategy. This approach splits a long video into multiple segments, which are encoded individually with overlapping frames to maintain good temporal consistency.
With the aid of this advanced video VAE, our xGen-VideoSyn-1 model is able to generate videos with over 100 frames at 720p resolution in an end-to-end manner. Figure \ref{fig:result-1} shows some example videos generated with our model.

In terms of the diffusion stage, we adopt a video diffusion transformer (VDiT) model that is architecturally similar to Latte~\cite{ma2024latte} and Open-Sora~\cite{opensora}.
Our VDiT incorporates transformer blocks with both temporal and spatial self-attention layers. We use ROPE~\cite{su2024roformer} and sinusoidal~\cite{vaswani2017attention} encodings for spatial and temporal position information. This allows for effective generalization across different lengths, aspect ratios, and resolutions. Moreover, our DiT model is trained on a diverse dataset that includes 240p, 512$\times$512, 480p, 720p, and 1024$\times$1024 resolutions. The video VAE training takes approximately 40 H100 days, while the DiT model requires around 642 H100 days.

Training data is crucial for text-to-video models. These models require high-quality video-text pairs to learn how to map text to video modalities. To address this, we designed a data processing pipeline that yields a large quantity of high-quality video-text pairs.
Our process includes removing duplicate data, analysis of aesthetics and motion, optical character recognition (OCR), and other processing steps. The process also captions videos. We developed a video captioning model that creates captions with an average of 84.4 words.
Our process is automatic and can be scaled as needed.
With this process, we have created a training dataset with over 13 million high-quality video-text pairs.

The contributions of our xGen-VideoSyn-1 framework can be summarized as follows:
\textbf{(1)}  We propose a novel video compression method that encodes long videos into latents with significantly reduced sizes.
\textbf{(2)} We have developed an automated data processing pipeline and created a large training set containing over 13 million high-quality video-text pairs.
\textbf{(3)}  xGen-VideoSyn-1 supports video generation with various sizes, durations, and aspect ratios, producing up to 14 seconds of 720p video.
\textbf{(4)}  Our model achieves competitive performance in text-to-video generation compared to state-of-the-art models.

\begin{figure}[t]
\centering
    \includegraphics[width=0.95\textwidth]{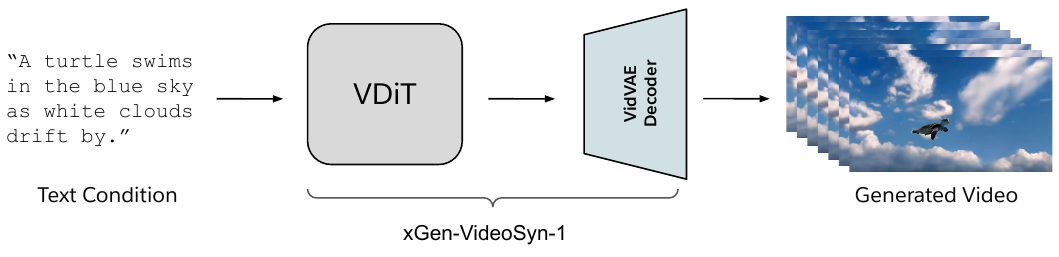}
    \caption{The core of our xGen-VideoSyn-1 model is a Video DiT (VDiT) module and a new Video VAE (VidVAE) module. The key is that our Video VAE module is able to encode and compress long video sequences into a latent representation during training; as well as reconstruct and decode such latent representation into long and realistic video sequences during inference. 
    }\label{fig:vid-model-simple}
    \vspace{-2mm}
\end{figure}

\section{Related Work}
\subsection{Video Generation}
Video generation has gained significant popularity in recent years, drawing considerable attention. Building on advancements in diffusion models for image generation, these techniques have been adapted for video generation, particularly through the use of 3D U-Net architectures~\cite{ho2022video}. To address the challenge of generating high-resolution videos, cascaded architectures have proven effective. For example, Imagen Video~\cite{ho2022imagen} and Make-a-Video~\cite{singer2022make} employ multi-stage pipelines that integrate spatial and temporal super-resolution networks. However, training such multi-stage models poses considerable difficulties with many hyper-parameters to tune.

Inspired by the success of Latent Diffusion, Video LDM~\cite{blattmann2023align} adapts a similar approach to the video domain by employing a Variational Autoencoder (VAE) to encode videos into latent representations. Other models such as Stable Video Diffusion (SVD) \cite{blattmann2023stable}, Lavie \cite{wang2023lavie}, and ModelScope~\cite{wang2023modelscope} utilize a 3D U-Net architecture to model diffusion processes in latent spaces.

The Diffusion Transformer (DiT) has gained prominence for its multi-scale flexibility and scalability. It effectively addresses the limitations of U-Net models which are often constrained by fixed-size data generation due to the inherent constraints of convolutional operations in local feature learning. 
DiT also benefits from acceleration techniques borrowed from Large Language Models (LLMs), facilitating easier scaling. Latte~\cite{ma2024latte}, a pioneering method, extends DiT to the video domain with the introduction of a spatial-temporal transformer block. Sora~\cite{videoworldsimulators2024} employs DiT as its backbone, inspiring further developments in the field. Open-source projects like Open-Sora~\cite{opensora} and OpenSoraPlan~\cite{pku_yuan_lab_and_tuzhan_ai_etc_2024_10948109} have emerged as leading open-source projects, continuing to push the boundaries in this field.

\subsection{Variational Autoencoders} Variational Autoencoders (VAEs) \cite{kingma2013auto} have become a prominent tool for image encoding. Two main approaches are typically employed: encoding images into continuous latent spaces \cite{rombach2022high,kingma2013auto}, and incorporating quantization techniques to learn discrete latent representations, as in VQVAE \cite{van2017neural} and VQGAN \cite{esser2021taming}. Expanding the application of VAEs, recent research has delved into encoding videos, aiming to leverage these encoded representations in text-to-video generation models. VideoGPT \cite{yan2021videogpt} employs a variant of VQ-VAE using 3D convolutions and axial self-attention to learn downsampled, discrete latent representations of raw videos. MAGVIT \cite{yu2023magvit} introduces a new 3D-VQVAE architecture focused on temporal compression; and its successor, MAGVIT-v2 \cite{yu2023language}, further refines the video tokenization process by integrating a lookup-free quantizer and various enhancements to the tokenizer model.

Despite their innovations, many of these models are not open-sourced or described with sufficient detail, which often leaves the research community with knowledge gaps in terms of their implementation details. Conversely, recent contributions from Open-sora-plan \cite{pku_yuan_lab_and_tuzhan_ai_etc_2024_10948109} and Open-Sora \cite{opensora}, have notably opened their methodologies, providing access to both code and model checkpoints. \cite{pku_yuan_lab_and_tuzhan_ai_etc_2024_10948109} expands upon traditional 2D VAEs by transitioning 2D convolutional layers to 3D causal convolutions and integrating temporal compression layers following spatial compression. \cite{opensora} introduces a cascading framework, initially encoding with a 2D VAE followed by additional compression via a 3D VAE. During the decoding phase, this model reconstructs the data first through a 3D VAE decoder and then through a 2D VAE decoder, providing a unique approach to temporal and spatial data handling in videos.

\section{Model Architecture}
\begin{figure}[t]
\centering
    \includegraphics[width=\textwidth]{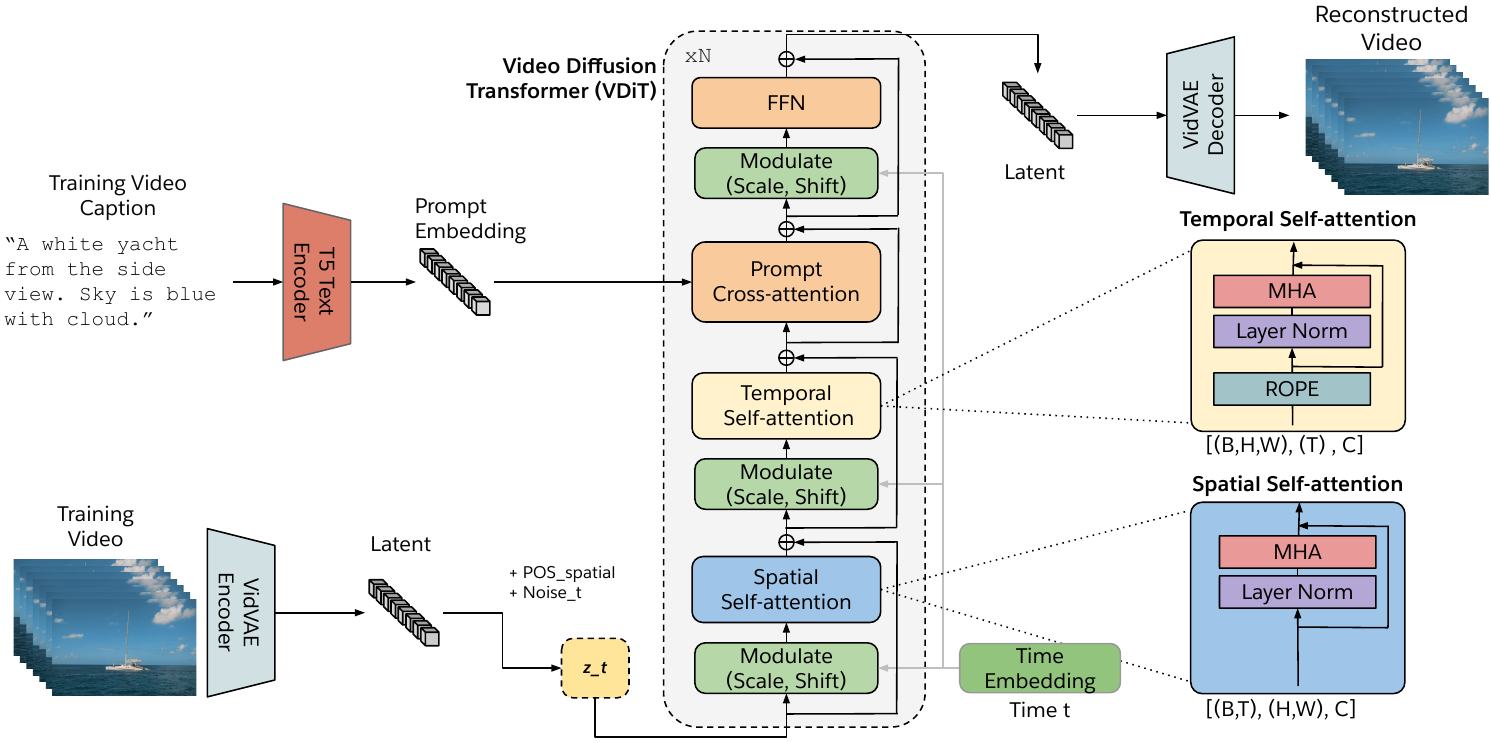}
    \caption{Detailed architecture of our proposed xGen-VideoSyn-1 model during training}\label{fig:arch}
    \vspace{-3mm}
\end{figure}
Our proposed xGen-VideoSyn-1 model comprises three main components: 1) a VideoVAE encoder and decoder, 2) a Video Diffusion Transformer (VDiT), and 3) a Language Model (Text Encoder). Further details are described below.

\subsection{VideoVAE}
\label{sec:videovae}

The task of the VideoVAE encoder is to take an input video and produce a latent encoding that can be used for reconstruction later. Our primary objective is to efficiently compress videos not only in the spatial dimension but also temporally, thereby enhancing training speed and reducing computation costs. Drawing inspiration from \cite{pku_yuan_lab_and_tuzhan_ai_etc_2024_10948109}, we enhance the conventional 2D VAE---used predominantly for spatial compression of still images---into a 3D variant capable of temporal compression by incorporating time-compacting layers. Originally introduced by \cite{kingma2013auto}, VAEs have been extensively utilized for image autoencoding, encoding an image into a latent feature space and subsequently reconstructing it from that space. Specifically, given an image $x \in \mathbf{R}^{H \times W \times 3}$ in RGB format, the encoder $\mathcal{E}$ maps $x$ to a latent representation $z = \mathcal{E}(x)$, and the decoder $\mathcal{D}$ reconstructs the image from $z$, such that $\tilde{x} = \mathcal{D}(z) = \mathcal{D}(\mathcal{E}(x))$, where $z \in \mathbf{R}^{h \times w \times c}$. The encoder reduces the dimensionality in the feature space by a factor of $f = H/h = W/w$.  To construct a 3D VideoVAE, we adapt a pre-trained 2D image VAE\footnote{\url{https://huggingface.co/stabilityai/sd-vae-ft-mse-original/blob/main/vae-ft-mse-840000-ema-pruned.ckpt}},
with a spatial compression rate of 1/8 from \cite{rombach2022high}. This adaptation involves the incorporation of time compression layers into the model. Similarly, for a video $x \in \mathbf{R}^{T \times H \times W \times 3}$, where $T$ represents the number of frames, the VideoVAE encodes $x$ into $z = \mathcal{E}(x)$, and $\mathcal{D}$ reconstructs the video from $z$, rendering $\tilde{x} = \mathcal{D}(z) = \mathcal{D}(\mathcal{E}(x))$, where $z \in \mathbf{R}^{t \times h \times w \times c}$. The encoder not only reduces the spatial dimensionality by a factor of $f = H/h = W/w$ but also compresses temporally by a factor of $s = T/t$. In our experiments, we achieve a temporal compression of 1/4.

To extend the 2D image-based VAE into a 3D VideoVAE, we implemented a series of modifications:
1) We replaced all 2D convolutional layers (Conv2d) with Causal Convolutional 3D layers (CausalConv3D). We opted for CausalConv3D to ensure that only subsequent frames have access to information from previous frames, thereby preserving the temporal directionality from past to future.
2) We introduced a time downsampling layer following the spatial downsampling layers to compact the video data along the temporal dimension. For this purpose, we utilized a 3D average pooling technique. Specifically, we incorporated two temporal downsampling layers, each reducing the temporal resolution by half. Consequently, the overall time compression factor achieved is 1/4, meaning that every four frames are condensed into a single latent representation. The spatial compression ratio remains 1/8.

\begin{figure}[t]
\centering
    \includegraphics[width=\textwidth]{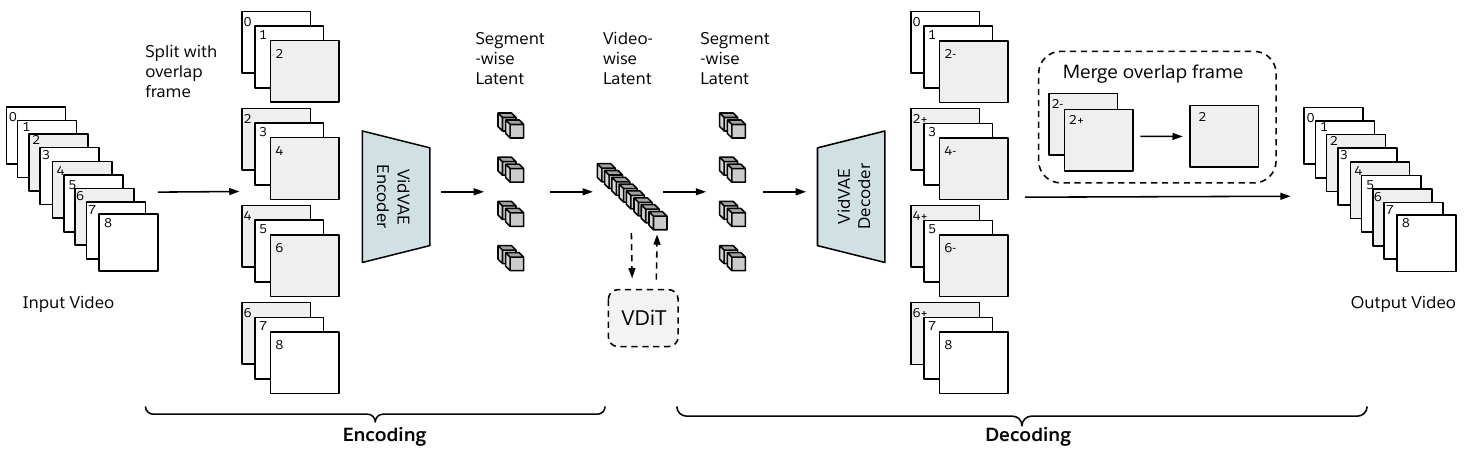}
    \caption{Video latent extraction pipeline}\label{fig:videovae}
    \vspace{-2mm}
\end{figure}

Despite achieving a 4$\times$8$\times$8 compression, the computation cost remains a significant bottleneck, particularly as video sizes increase, leading to substantial memory demands. To address the out-of-memory (OOM) issues encountered during long video encoding, we propose a divide-and-merge strategy. As illustrated in Figure \ref{fig:videovae}, this approach involves splitting a long video into multiple segments. Each segment consists of five frames, with duplicate frames at both the beginning and end. These segments are encoded individually, using overlapping frames to maintain strong temporal consistency. With this video Variational Autoencoder (VAE) framework, our xGen-VideoSyn-1 model can generate over 100 frames of 720p video in an end-to-end manner, while mitigating additional computation costs.

\subsection{Video Diffusion Transformer (VDiT)} 
Our Video Diffusion Transformer is based on the architecture of Open-Sora~\cite{opensora} and Latte~\cite{ma2024latte}, utilizing a stack of spatial-temporal transformer blocks as illustrated in Figure~\ref{fig:arch}. Each transformer module incorporates a pre-norm layer and multi-head self-attention (MHA).  We use Rotary Positional Embedding (RoPE)~\cite{su2024roformer} to encode temporal information and sinusoidal encoding~\cite{vaswani2017attention} for spatial information. For text feature extraction, we employ the T5 model~\cite{raffel2020exploring} with a token length limit of 250. The extracted text features are integrated into the backbone through a cross-attention layer. We follow the PixArt-Alpha~\cite{chen2023pixart} model to encode the time-step embedding, incorporating a modulation layer within each transformer block.

We take the latent diffusion model (LDM) for training~\cite{rombach2022high}. It follows the standard DDPM~\cite{ho2020denoising} with denoising loss and uses Diffusion Transformer (DiT)~\cite{peebles2023scalable} as the diffusion backbone. To enable generative controllability, our model has applied conditioning caption signals ($y$), encoded aside by language model T5 and injected into the DiT, with the help of cross-attention layers. This can be formulated as:
\begin{equation}
    \mathcal{L}_{LDM} := \mathbb{E}_{z\sim E(x), \varepsilon \sim N(0,1), t, y} \left [ \left \| \varepsilon - \varepsilon_{\theta}(z_t,t,c_{\phi}(y)) \right \| \right ],
\end{equation}
where $t$ represents the time step, $z_t$ is the noise corrupted latent tensor at time step $t$,
and $z_0 = \mathcal{G}(x)$. $\varepsilon$ is the unscaled Gaussian noise,
$c_{\phi}$ is the conditioning network parameterized by $\phi$ and $\varepsilon_{\theta}$ is the Transformer-like denoising network (video decoder).
The parameters of both conditioning and denoising networks ${\theta, \phi}$, are trained by the LDM loss. During inference,  clean videos can be generated via  classifier-free guidance~\cite{ho2022classifier} as:
\begin{equation}
\hat\varepsilon_{\theta}(z_t|y) = \varepsilon_{\theta}(z_t) + s \cdot (\varepsilon_{\theta}(z_t, c_{\phi}(y)) - \varepsilon_{\theta}(z_t)),
\label{eq:sampling}
\end{equation}
where $s$ is the guidance weight to balance text controllability and image fidelity.

\section{Training Data Collection and Processing} 

\begin{figure}[t]
\centering
    \includegraphics[width=\textwidth]{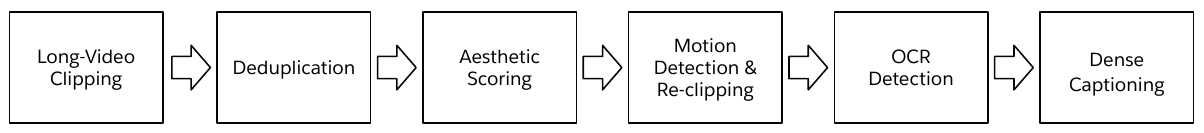}
    \caption{Training data collection and processing pipeline}\label{fig:data-processing}
    \vspace{-2mm}
\end{figure}

The pipeline operates sequentially, as shown in Figure~\ref{fig:data-processing}. First, it splits long videos into manageable clips. Then, it removes similar and redundant clips. Next, it analyzes aesthetics and motion dynamics across frames to eliminate static video clips and inconsistent frames. After that, it identifies and removes clips contaminated with text or watermarks. Finally, it evaluates and scores the visual quality of clips before adding descriptive captions to the clips. We further describe these steps in this section.

\subsection{Video Clipping}
Original long videos are cut into multiple shorter clips using the PySceneDetect\footnote{\url{https://github.com/Breakthrough/PySceneDetect}} tool. Each clip is intended to represent a distinct and clean scene. However, some clips may still contain redundant or inconsistent scenes. These cases are addressed in subsequent steps.

\subsection{Deduplication}
The clipping process can sometimes yield clips that are highly similar to one another. To address this, a de-duplication step is essential to filter out redundant clips. We use ffmpeg\footnote{\url{https://ffmpeg.org}} to extract frames and the clip-as-a-service tool\footnote{\url{https://github.com/jina-ai/clip-as-service/tree/main}} to efficiently extract CLIP features and compute similarity scores between clips. In each duplicate pair, we remove the shorter clip based on a similarity score threshold,  $\tau$. Through empirical analysis, we have found that a threshold of  $\tau$ = 0.9 is effective for identifying duplicates.

\subsection{Aesthetic Scoring}
To ensure high-quality training data, it is crucial to use video clips that are well-lit, well-composed, and have clear footage. To filter out poor-quality data, we compute the Aesthetic Score—a measure of how visually pleasing a video is. We utilize a simple neural network\footnote{\url{https://github.com/christophschuhmann/improved-aesthetic-predictor/tree/main}} trained on human aesthetic scores of images. This network, which takes CLIP features as input, outputs a score ranging from 0 to 10. Clips with an Aesthetic Score below 4.5 are filtered out.

\subsection{Motion Detection and Re-clipping} 
In this video processing step, we aim to achieve two primary goals. Firstly, we want to eliminate videos that are nearly static. Secondly, after the initial video clipping, some videos may still exhibit sudden scene changes. For these videos, we will re-clip them to ensure consistency and maintain a unified topic throughout. Our approach utilizes frame differencing to detect motion within a video, followed by motion-based re-clipping. The process commences with the computation of grayscale frame differences, where we subtract each frame from its predecessor in the sequence. This technique, while effective, can introduce background noise, manifesting as speckles that falsely indicate motion. These artifacts typically stem from minor camera shakes or the presence of multiple shadows. To counteract this, we implement a threshold on the frame differences to create a binary motion mask. To refine the quality of this motion mask, we apply techniques such as blurring \cite{culjak2012brief} and morphological operations \cite{comer1999morphological}. Following this, we calculate a motion score by taking the mean of the motion mask values.

\begin{figure}[t]
    \centering
    \includegraphics[width=0.98\linewidth]{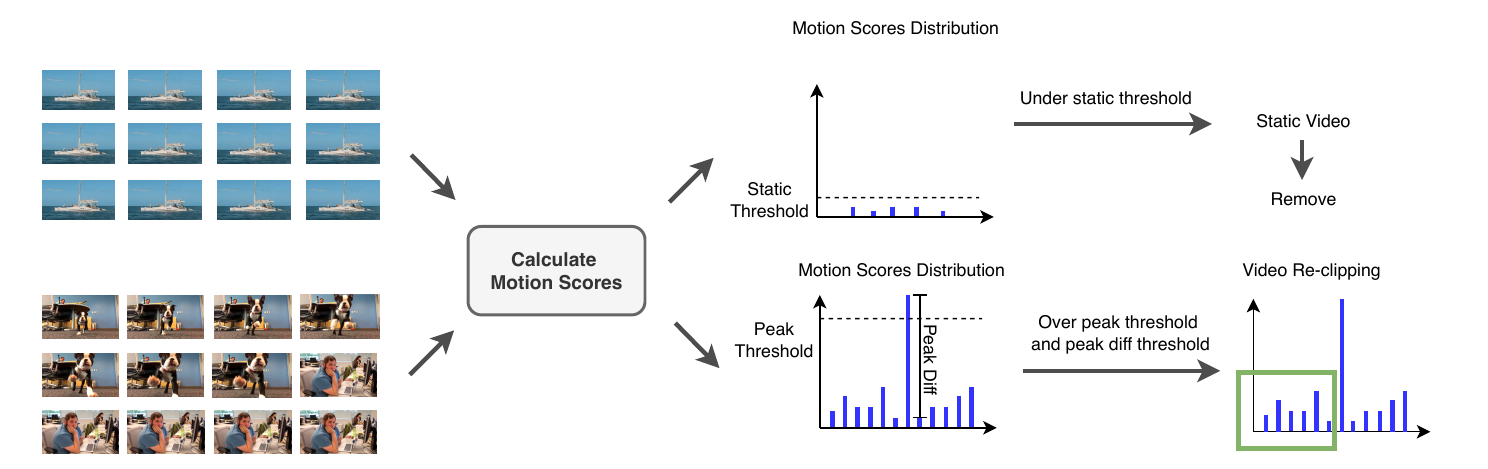}
    \caption{Process of motion detection and re-clipping}
    \label{fig:motion_detection}
\end{figure}

Guided by the motion score, we perform both motion detection and re-clipping. An overall illustration is shown in Figure~\ref{fig:motion_detection}. We calculate the average motion score across the video and set a threshold. The overall distribution of the average motion score is illustrated in Figure~\ref{fig:motion_distribution} of the Appendix. Videos falling below this threshold are deemed nearly static and subsequently removed. For the re-clipping, our criteria focus on eliminating significant, sudden scene changes. We identify the frame with the highest motion score and analyze the motion score differences with its neighboring frames. If both the peak motion score and the differences surpass predefined thresholds, this flags a major scene change. Here, we segment the video at this critical frame. We retain the longer segment, ensuring it meets the length requirement and is devoid of further disruptive transitions.

\begin{table*}[t]
    \centering
    \small
    \setlength\tabcolsep{4.8pt}
  \vspace{-3mm}
  \scalebox{1.00}{
  \begin{tabular}{lccccccc}
    \toprule
    Dataset &  Domain & \#Videos & \multicolumn{2}{c}{Avg/Total Video len} & Avg caption len & Resolution \\
    \midrule
    MSVD~\cite{msvd}            & Open    & 1970   & 9.7s  & 5.3h     & 8.7 words  & -     \\
    LSMDC~\cite{lsmdc}            & Movie   & 118K   & 4.8s  & 158h     & 7.0 words  & 1080p \\
    MSR-VTT~\cite{msrvtt}         & Open    & 10K    & 15.0s & 40h      & 9.3 words  & 240p  \\
    DiDeMo~\cite{didemo}             & Flickr  & 27K    & 6.9s  & 87h      & 8.0 words  & -     \\
    ActivityNet~\cite{activitynet}     & Action  & 100K   & 36.0s & 849h     & 13.5 words & -     \\
    YouCook2~\cite{youcook2}     & Cooking & 14K    & 19.6s & 176h     & 8.8 words  & -     \\
    VATEX~\cite{vatex}             & Open    & 41K    & $\sim$10s & $\sim$115h & 15.2 words & - \\
    WebVid-10M~\cite{Bain21} &  Open   & 10M  & 18s  & 52k h & 12.0 words &  336p\\
 
    Panda-70M~\cite{chen2024panda70m}     & Open    & 70.8M  & 8.5s  & 166.8Khr & 13.2 words & 720p \\
    OpenVid-1M~\cite{nan2024openvid}     &  Open   & 1M  &  - & - & Long & 720p-1080p \\
    \midrule
    xGen-VideoSyn-1    & Open    &  13M & 6.9s & 25K h  & 84.4 words & 720p-1080p \\
    \bottomrule
    \end{tabular}}
    \caption{Comparison of our dataset and other video-language datasets\label{tab:dataset}
    }
    \vspace{-3.5mm}
\end{table*}

\subsection{Optical Character Recognition (OCR)}

We also conduct OCR to detect text in the video in order to get high quality video data. The tool we used is PaddleOCR\footnote{\url{https://github.com/PaddlePaddle/PaddleOCR}}. We performed text detection on key frames from the videos. The text detection model we used is ``ch\_PP\-OCRv4\_det\_infer'', a lightweight model supporting Chinese, English, and multilingual text detection. In this step, we only kept videos where the size of the bounding box is smaller than 20000 pixels.

\subsection{Dense Captioning}

We train a multimodal video LLM to generate video captions. This model takes a sequence of frames from the video as an input, and is trained to generate text captions describing the contents of the video as an output.

\begin{figure*}[t]
  \centering
  \begin{subfigure}[b]{0.42\linewidth}
    \includegraphics[width=\linewidth]{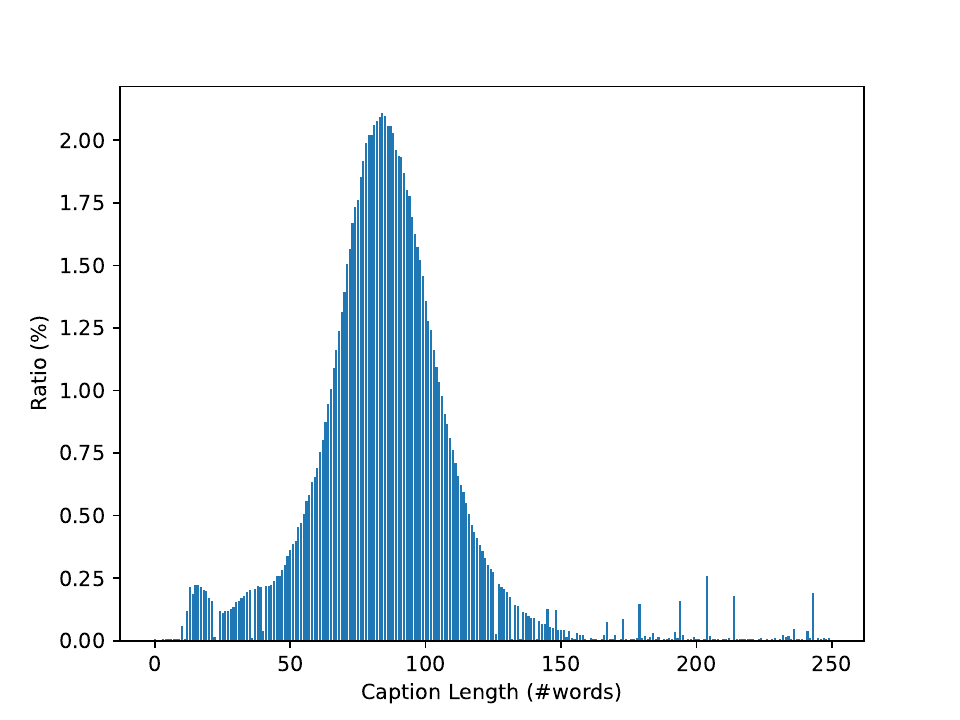}
    \caption{Distribution of caption length}
  \end{subfigure}
    \begin{subfigure}[b]{0.57\linewidth}
    \includegraphics[width=0.8\linewidth]{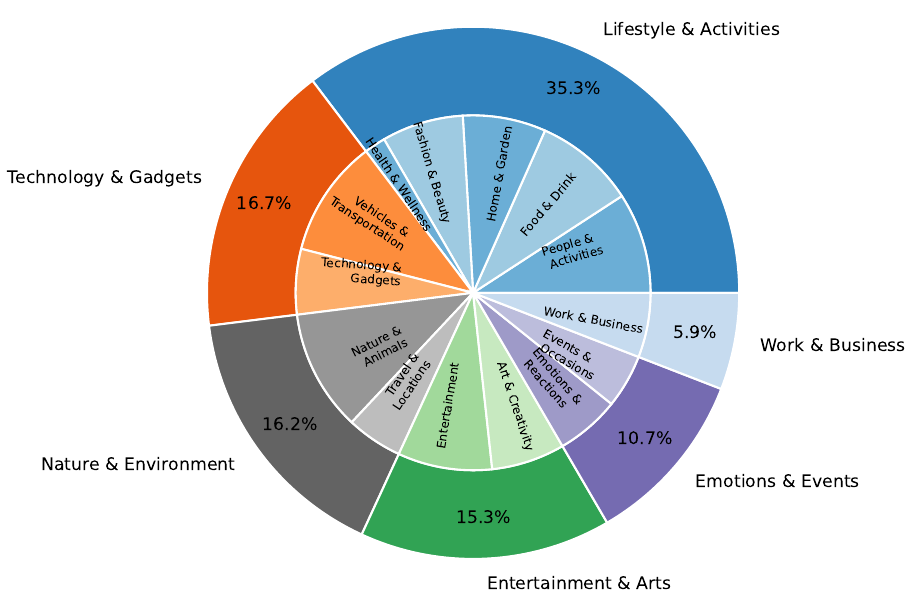}
    \caption{Category distribution}
  \end{subfigure}
   \caption{Caption statistics analysis\label{fig:caption_analysis}}
   \end{figure*}

   \begin{table}[t]
    \centering
    \small
    \scalebox{0.85}{
    \begin{tabular}{c|cccccc}
    
    \hline
        Methods & \#Params  & GPU Days  & Data & VAE  & Max Resolution & Max Duration \\
        \hline
        OpenSoraPlan V1.1 &1.0B  &240 (H100) + 1536 (Ascend)  &4.8M  &4x8x8  & 512$\times$512 &9.2s \\
        OpenSoraPlan V1.2 &2.77B  &1578 (H100) + 500 (Ascend)  &6.1M  &4x8x8  & 720p &4s \\
        OpenSora V1.1 &700M  &576 (H800)  &10M  &1x8x8  & 720p  &4s \\
        OpenSora V1.2 &1.1B  &1458 (H100)  & >30M &4x8x8  & 720p &16s \\
        Ours  &731M  &672 (H100)  &13M  &4x8x8  & 720p &14s \\
        \hline
    \end{tabular}}
    \caption{Settings of different text-to-video models }
    \label{tab:setting}
    \vspace{-3mm}
\end{table}

\noindent\textbf{Captioning Model.} Our video captioning model is an extended version of xGen-MM~\cite{Xue2024xGenMMA}. The model architecture is composed of the following four components: (1) a vision encoder (ViT) taking each frame input, (2) a frame-level tokenizer to reduce the number of tokens, (3) a temporal encoder to build video-level token representations, and (4) a LLM generating output text captions based on such video tokens and text prompt tokens.

Specifically, we use a pretrained ViT-H-14 \cite{dosovitskiy2020vit} as the vision encoder, designed to take one single image frame at a time. Perceiver-Resampler \cite{alayrac2022flamingo} is then applied to map such visual tokens into $N=128$ visual tokens per frame. The temporal encoder is implemented with Token Turing Machines (TTM) \cite{ryoo2023tokenturingmachines}, which is a sequential model capable of taking any number of frames to generate a video-level token representation (e.g., $M=128$ tokens regardless the number of frames). Our use of TTM is similar to its usage in Mirasol3B \cite{piergiovanni2024mirasol3b}, except that our model uses TTM directly to encode a sequence of image tokens while Mirasol3B uses TTM to encode a sequence of low-level video tokens. We use Phi-3 \cite{abdin2024phi3} as our multimodal LLM taking such video tokens in addition to the text prompt tokens. For computational efficiency, the model takes uniformly sampled 4 frames per video. Our model uses ViT to map a video into around $4 \times 700$ visual tokens. These visual tokens are then mapped to $4 \times 128$ visual tokens using Perceiver-Resampler and then to 128 video tokens using TTM.

The model is first pretrained with standard image caption datasets. The model is then finetuned with the LLaVA-Hound-DPO training dataset \cite{zhang2024direct}, providing video captions over 900k frames. Instead of directly using the text captions provided in LLaVA-Hound-DPO, we used Mistral-8x7B~\cite{jiang2024mixtral} to rephrase such text captions so that they become more Sora-style captions.

We use a very straight forward text prompt input to generate the captions: `A chat between a curious user and an artificial intelligence assistant. ``The assistant gives helpful, detailed, and polite answers to the user's questions.'' Please provide a description of this video.'

\noindent\textbf{Analysis of Captioning Results.}
We randomly sampled 100k captions. Figure~\ref{fig:caption_analysis} (a) shows the caption length distribution and work cloud for these sampled captions. The average caption length is 84.4 words, which is much longer than other video-language datasets as we know. Additionally, about 87\% of captions range from 50 to 120 words.

\begin{figure}[t]
    \centering
    \includegraphics[width=1.0\linewidth]{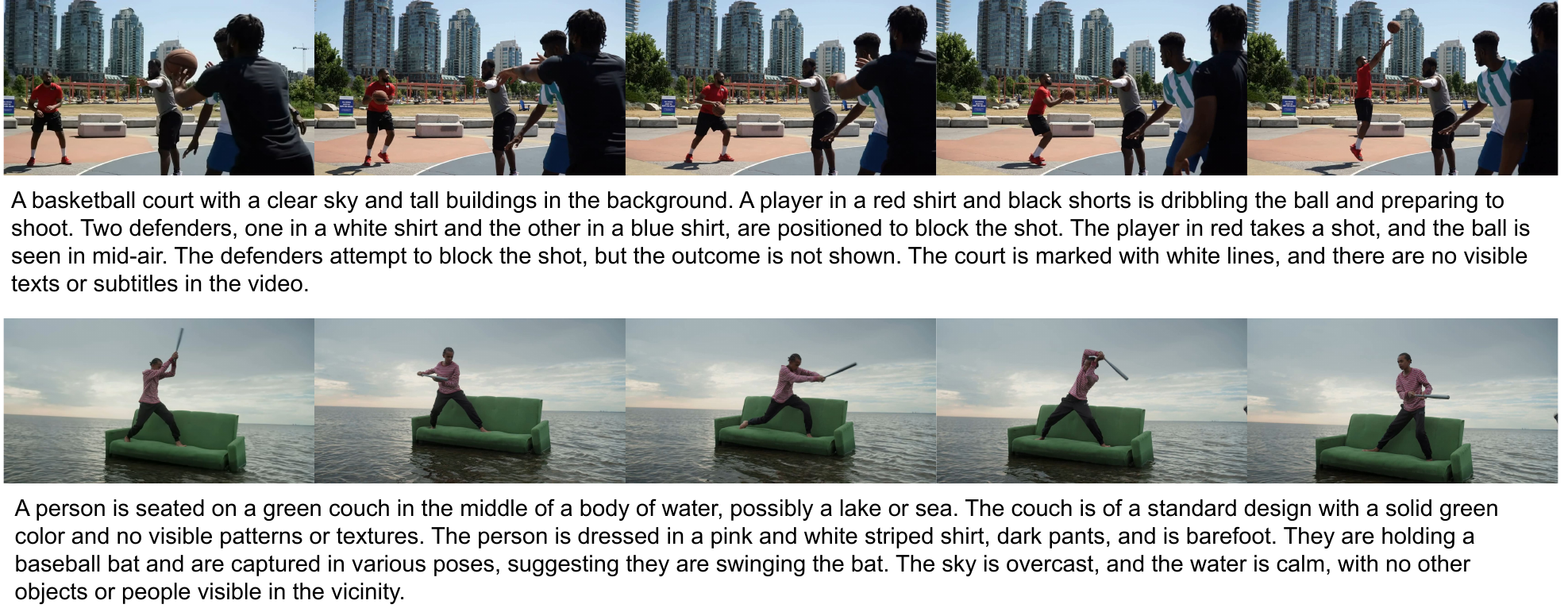}
    \caption{Example text captions generated using our xGen-MM-based video Captioner. We use such generated video-text pairs for the video generation training.}
    \label{fig:caption_examples}
\end{figure}

\begin{figure}[t]
    \centering
    \includegraphics[width=0.5\linewidth]{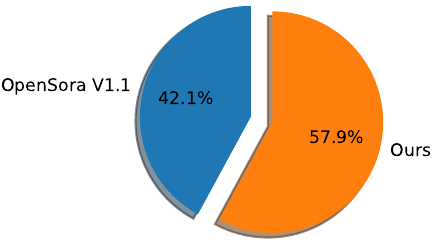}
    \caption{User study of text-to-video generation}
    \label{fig:user_study}
\end{figure}

\begin{table}[t]
    \centering
    \small
    \scalebox{0.9}{
    \begin{tabular}{c|ccc|ccc|ccc}
    \hline
         \multirow{2}{*}{Methods} &\multicolumn{3}{|c|}{1$\times$768$\times$768}  &\multicolumn{3}{|c|}{17$\times$512$\times$512}  &\multicolumn{3}{|c}{65$\times$256$\times$256} \\
         
         &PSNR$\uparrow$ &SSIM$\uparrow$  &MSE$\downarrow$  &PSNR$\uparrow$ &SSIM$\uparrow$ &MSE$\downarrow$  &PSNR$\uparrow$ &SSIM$\uparrow$  &MSE$\downarrow$ \\
         \hline
         ImageVAE~\cite{rombach2022high} &40.98  &0.972 &0.00067  &37.59  &0.951 &0.00152  &32.54  &0.901 &0.00472   \\
         \hdashline
         OpenSoraPlan~\cite{pku_yuan_lab_and_tuzhan_ai_etc_2024_10948109} &39.15  &\tbf{0.973}  &\tbf{0.00082}  &33.62  &0.934  &0.00289  &\tbf{30.06}  &0.874  &0.00814    \\
         Ours&\tbf{39.41}  &0.971  &\tbf{0.00082}  &\tbf{33.83}  &\tbf{0.935}  &\tbf{0.00281}  &29.68  & \tbf{0.879}  &\tbf{0.00780}  \\
         
         \hline
    \end{tabular}}
    \caption{VideoVAE quantitative evaluation}
    \label{tab:vidvae}
\end{table}
   
We pre-defined six scene-specific categories for videos: ``Lifestyle \& Activities'', ``Nature \& Environment'', ``Technology \& Gadgets'', ``Entertainment \& Arts'', ``Work \& Business'', and ``Emotions \& Events'', along with their respective subcategories. For each video caption,  we asked OpenAI's fastest model ``gpt-4o-mini'' to select the most appropriate subcategory. Figure~\ref{fig:caption_analysis} (b) displays the category distribution. The video captions contain diverse of categories, making it a value resource for video generation. Figure~\ref{fig:caption_word_cloud} shows the word cloud of our captions.

\subsection{Distributed Data Processing Pipeline}
To efficiently orchestrate the six data processing and filtering steps described above with minimal manual intervention and optimal resource utilization, we employ a Distributed Data Processing Pipeline. We use RabbitMQ to manage a video processing pipeline. Each stage is deployed with specific resources and processes videos through multiple queues. The pipeline starts by adding videos to the initial queue, with each subsequent stage filtering and processing clips based on predefined criteria. Further details can be found in the Section~\ref{sec:distributed-data} of Appendix.

\section{Evaluation}
We empirically evaluate the effectiveness of xGen-VideoSyn-1 through a series of comprehensive experiments across various tasks, including video generation and compression. Details on the experimental setup, methodologies, and results analysis are provided in the following sections.

\subsection{Implementation Details}
Our proposed xGen-VideoSyn-1 model integrates a 731M diffusion transformer with a 244M video VAE model, trained sequentially. See more details in Table \ref{tab:setting}.

The video VAE model can compress the video by 4x8x8. It is trained on a subset of the Kinetics~\cite{kay2017kinetics} dataset and additional high-quality internal videos. We sample multi-scale images and videos from the training set, including resolutions of 1×768×768, 17×512×512, and 65×256×256. The model, initialized with the image VAE, requires 40 H100 days to train. For further details, please refer to Section~\ref{sec:videovae}.

The video DiT model features 28 stacked transformer blocks, with each multi-head attention (MHA) layer consisting of 16 attention heads and a feature dimension of 1152. This DiT model encompasses 731 million parameters in total. We adopt a training pipeline similar to OpenSora V1.1, utilizing multiple buckets to accommodate various sizes, aspect ratios, and durations. The DiT model is initialized using the PixArt-Alpha~\cite{chen2023pixart} model and undergoes training in three stages: the first stage with videos up to 240p, the second stage with videos up to 480p, and the third stage with videos up to 720p. We use AdamW with a default learning rate of 2e-5, and the final checkpoint is obtained through exponential moving average (EMA). The overall training process spans approximately 672 H100 days. This DiT model can support up to 14s 720p video generation.

\begin{figure*}[t]
\centering
    \includegraphics[width=1.0\textwidth]{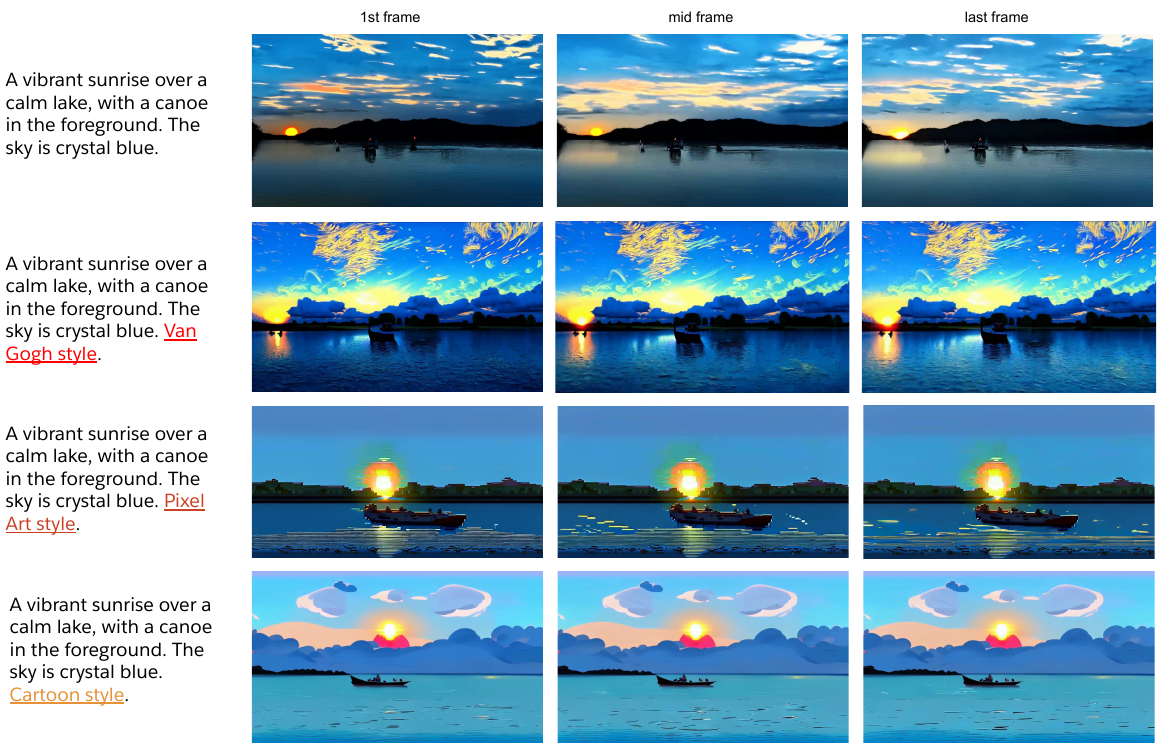}
    \caption{Prompt-based style control }\label{fig:style-result}
\end{figure*}

\subsection{Text-to-Video Generation}
\subsubsection{Quantitative Results}
We use Vbench~\cite{huang2024vbench} to quantitatively evaluate the text-to-video generation results. Tab.~\ref{tab:vbench} presents various scores for comprehensive evaluation. These scores are categorized into the following metrics: “Consistency” (including Background Consistency, Subject Consistency, and Overall Consistency), “Aesthetic” (including Aesthetic, Image Quality, and Color), “Temporal” (including Temporal Flickering, Motion Smoothness, and Human Action), and “Spatial” (spatial relationship). OpenSora V1.1, which is comparable to our model in size ($\sim$700M) and training cost, provides a fair benchmark. The ModelScope~\cite{wang2023modelscope} represents a Stable Diffusion-based method. We conduct the evaluation of OpenSora V1.1 and Ours under the same setting. ModelScope's scores are referred to the official table. As shown in Tab.~\ref{tab:vbench}, our model outperforms the baselines in “Aesthetic,” “Spatial”, and average results, while performing comparably to the baselines in other metrics.

\subsubsection{User Study}
We also conducted a user study on 2s 720p text-to-video generation to evaluate text controllability, as shown in Figure~\ref{fig:user_study}, using Amazon Mechanical Turk~\cite{doi:10.1177/1745691610393980}. Approximately 100 prompts, each around 55 words in length and randomly generated by ChatGPT, were used to cover a wide range of scenarios and challenging cases. The percentages of user votes for the three methods are reported in Figure~\ref{fig:user_study}. In this study, our model outperformed the baseline by more than 15\%, indicating a significant improvement. Additionally, the p-value, computed as 0.03 with three repetitions of the user study, is statistically significant with a threshold of $<$0.05.

\subsubsection{Style Control}
To demonstrate the capacity of our model in content creation, we conducted an ablation study on prompts related to style control. As illustrated in Figure~\ref{fig:style-result}, we applied a sample prompt with various styles. Our model successfully interprets and generates content in the desired styles, including “Van Gogh”, “Pixel Art”, and “Cartoon”. By default, the style tends to be realistic. However, applying style control prompts can sometimes reduce the prominence of other elements, such as sunrise, as observed with static sun imagery in the latter two rows. This highlights a limitation of our current model, which may be mitigated by scaling up the model size.

\subsection{Video Compression}
To further assess the reconstruction capacity of our trained video VAE, we randomly sampled 1,000 videos from the Kinetics~\cite{kay2017kinetics} and OpenVid1M~\cite{nan2024openvid} datasets, ensuring these videos were not included in the training set. We then used the VAE model to encode and decode these videos, expecting the outputs to be identical to the inputs. We evaluated the results using PSNR, SSIM~\cite{5596999}, and mean squared error (MSE) metrics. As shown in Tab.~\ref{tab:vidvae}, our model outperforms the baseline video VAE from OpenSoraPlan, which has the same compression ratio of 4x8x8, in most scenarios. Nevertheless, there remains a significant gap between the image VAE and our video VAE, indicating substantial potential for future improvements. The image VAE cannot compress videos at the time dimension which leaves huge redundancy in computation.

\begin{table}[t]
    \centering
    \scalebox{1.0}{
    \begin{tabular}{c|ccccc}
     \hline
        Methods & Consistency & Temporal   &Aesthetic   & Spatial & Avg \\
         \hline
         ModelScope~\cite{wang2023modelscope} &0.702 &\tbf{0.955}  &0.641  &0.337 &0.659 \\
        OpenSora V1.1~\cite{opensora} &\tbf{0.716}  &0.941  &0.599  &0.520 &0.694 \\
        Ours &0.714  &0.947  &\tbf{0.655}  &\tbf{0.523} &\tbf{0.709} \\
         \hline
    \end{tabular}}
    \caption{Vbench T2V score}
    \label{tab:vbench}
\end{table}

\section{Conclusion}
This work explores the architecture and technologies of the T2V model, focusing on the integration of video VAE and Diffusion Transformer (DiT) architectures. Unlike existing models that utilize image VAEs, our approach incorporates a video VAE to enhance both spatial and temporal compression, addressing the challenges of long token sequences. We introduce a divide-and-merge strategy to manage out-of-memory issues, enabling efficient encoding of extended video sequences. Our xGen-VideoSyn-1 model supports over 100 frames in 720p resolution, and the accompanying DiT model uses advanced encoding techniques for versatile video generation. A robust data pipeline for generating high-quality video-text pairs underpins our model's competitive performance in text-to-video generation.

%
%
\bibliographystyle{unsrtnat}
\bibliography{main}

\clearpage
\appendix

{
\Large
\textbf{Appendix}
}
\section{Distributed Data Processing Pipeline}
\label{sec:distributed-data}
\begin{figure}[h]
\centering
    \includegraphics[width=0.8\textwidth]{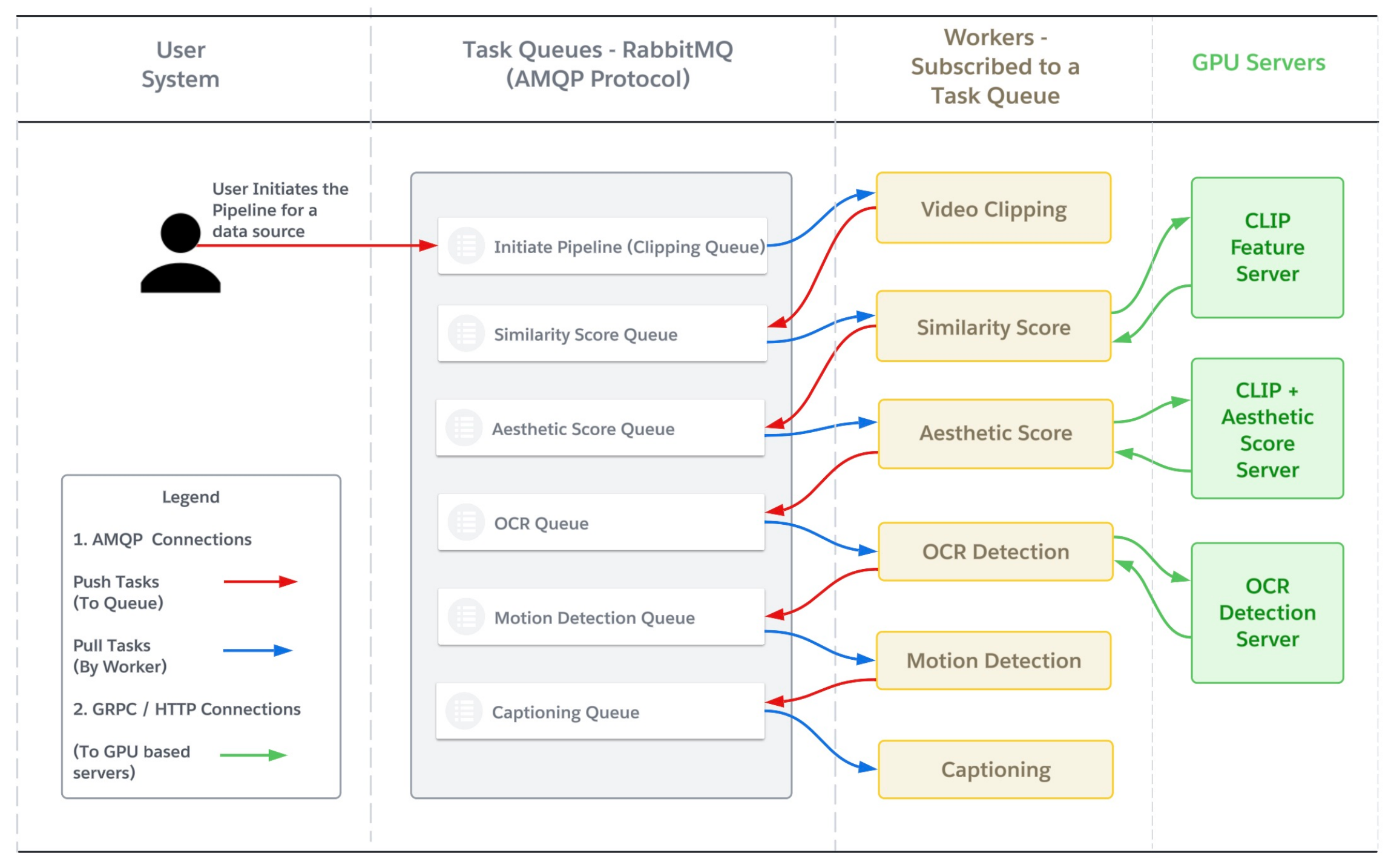}
    \caption{ System design of automatic high-quality video-text data collection pipeline}\label{fig:data-pipe}
\end{figure}
The Distributed Data Processing Pipeline meets the following criteria:
\begin{enumerate}
    \item Each process (one of the six steps above) is able to use its own resource specs. For example, clipping is CPU based, captioning is GPU based.
    \item Each process is independently scalable without interrupting the process flow. For example, Clipping is extremely fast, while Similarity Score is time consuming. Hence, we independently scale-down the clipping process and scale-up the Similarity Score process.
    \item The downstream processes are automatically triggered after a process is completed for a video. For example, after clipping is complete for video with ID `A', the similarity score computation is started for that video automatically.
    \item The activation of a downstream task optionally depends on a condition. For example, The motion detection for a clip is triggered only if the clip does not have a text, which is a result of the OCR detection process.
\end{enumerate}

To achieve the pipeline, we use RabbitMQ\footnote{\url{https://www.rabbitmq.com}} as a Task Orchestrator (Figure~\ref{fig:data-pipe}). Each process is a deployment with it's own resource specification, subscribed to a Task Queue. To trigger the pipeline, we start by pushing the video IDs to the initial queue. This is the only manual step required. Once this is done, the clipping process populates the Similarity Score Queue with the video ID. The Similarity Score deployement, subscribed to the corresponding queue, takes up the task, completes it, and pushes only the de-duplicated clip IDs to the Aesthetic Score queue. The Aesthetic score process, after computation, enqueues the OCR detection Queue with only the IDs of only those clips that meet the threshold. In this fashion, the number of clips that are being processed keeps reducing with each step in the pipeline by skipping the computation for clips that do not meet the passing criteria in the previous steps in the pipeline. In addition to the speed gain by skipping computation for failed clips, we also achieve the speed gain due to pipelining. The time taken for each step to process 1000 videos is given in Tab. \ref{tab:data-processing-time}. Equation \ref{eq:pipeline-efficiency} shows that the pipelined system is 1.5 times faster than the equivalent sequential system. There is scope to further improve this by speeding up the bottleneck process, as the processing time of the pipelined system is dependent on the time taken by the bottleneck process. As shown in Figure~\ref{fig:data-pipe}, our data collection pipeline includes multiple modules such as clipping, captioning and other process.

\begin{align}
    T_{Sequential} &= 1 + 3 + 0.8 + 1.2 + 12 \notag \\
    &= \text{18 minutes} \\
    T_{Pipelined}&~= \text{12 minutes (Time taken by bottleneck step)}\\
    efficiency &= \frac{T_{Sequential}}{T_{Pipelined}} \notag \\
    &= 1.5 \label{eq:pipeline-efficiency}
\end{align}
\begin{table}[]
    \centering
    \begin{tabular}{c|c|c}
    \hline
         Data Processing Step & Time taken for 1000 videos (in minutes) & Pass Rate (\%)\\
    \hline
         Clipping & 1 & N/A\\
         Deduplication & 3 & 64.18\\
         Aesthetic Score & 0.8 & 90.23\\
         OCR Detection & 1.2 & 67.9\\
         Motion Detection & 12 & 88.6\\
    \hline
    \end{tabular}
    \caption{Time taken and pass rate for each data processing step}
    \label{tab:data-processing-time}
\end{table}

\clearpage

\begin{figure}[tb]
    \centering
    \includegraphics[width=0.95\linewidth]{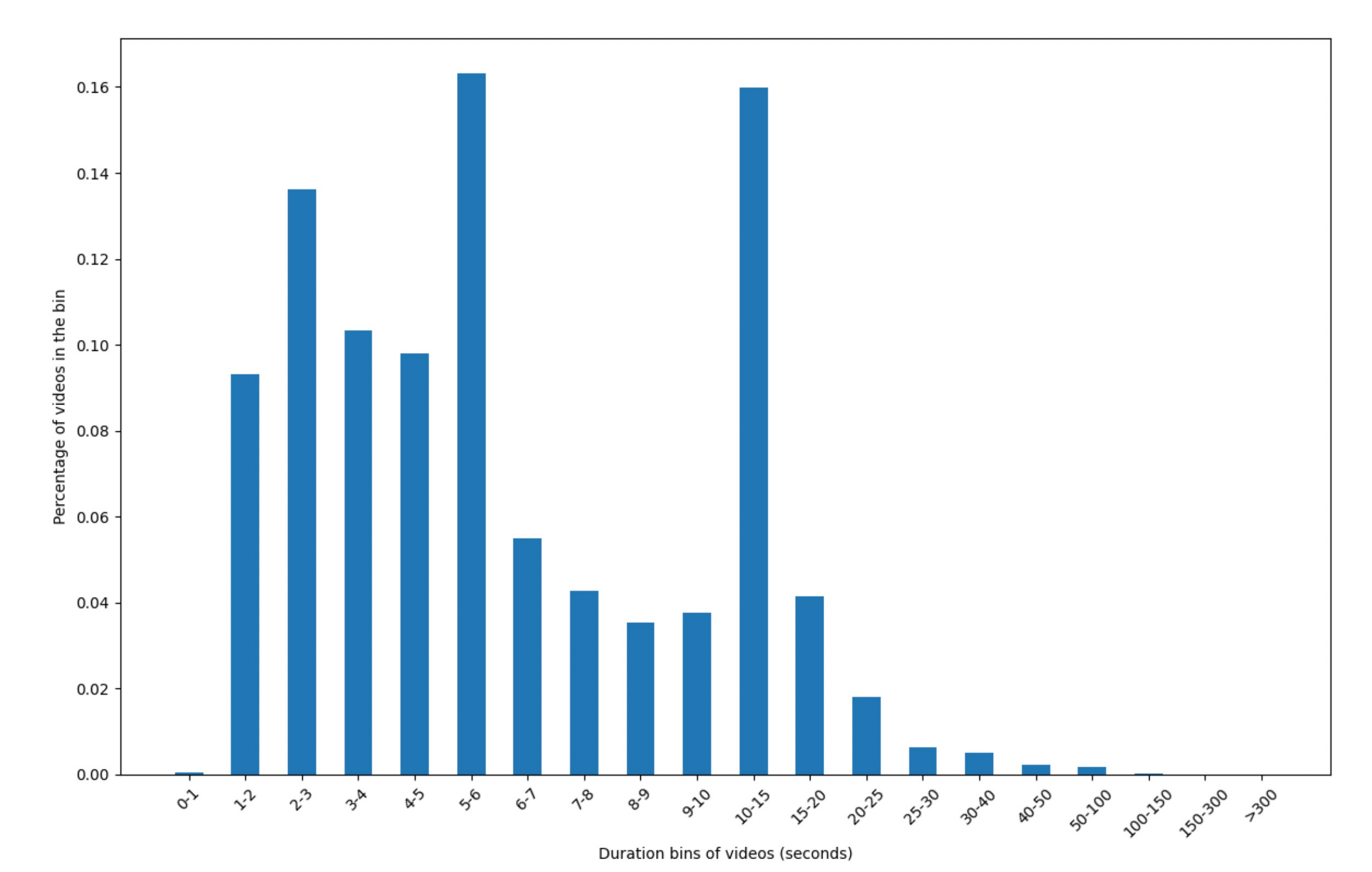}
    \caption{Distribution of video duration}
    \label{fig:duration-distribution}
\end{figure}

\begin{figure}[tb]
    \centering
    \includegraphics[trim={5cm 0.5cm 5cm 0.5cm}, clip, width=\linewidth]{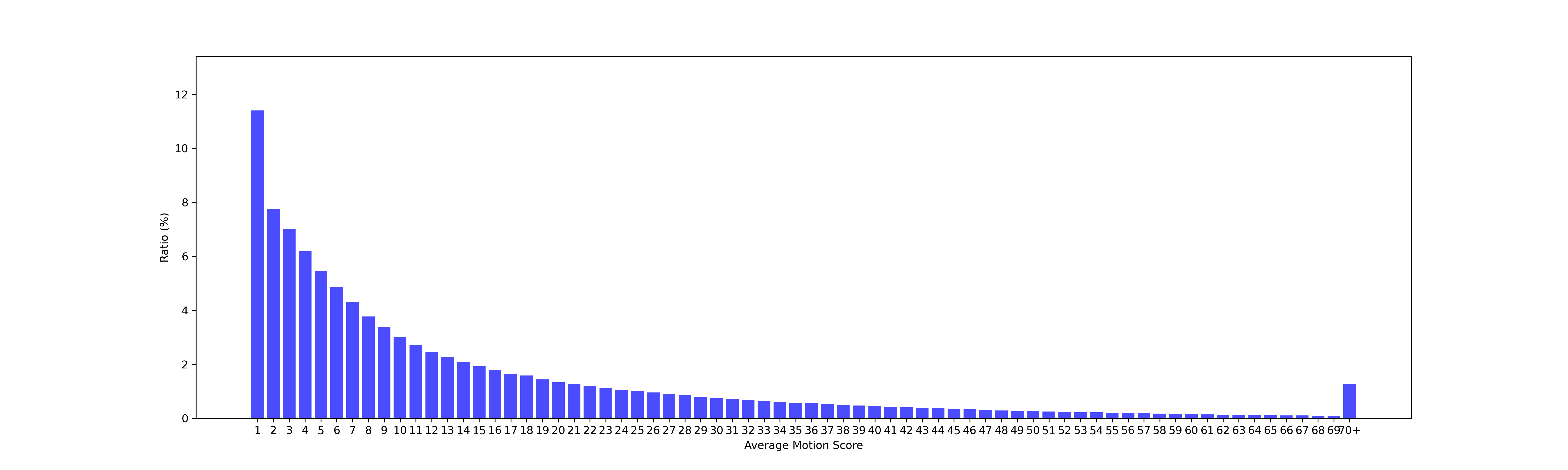}
    \caption{The overall distribution of average motion scores}
    \label{fig:motion_distribution}
\end{figure}

\clearpage

\begin{figure}[tb]
    \centering
    \includegraphics[width=1.0\linewidth]{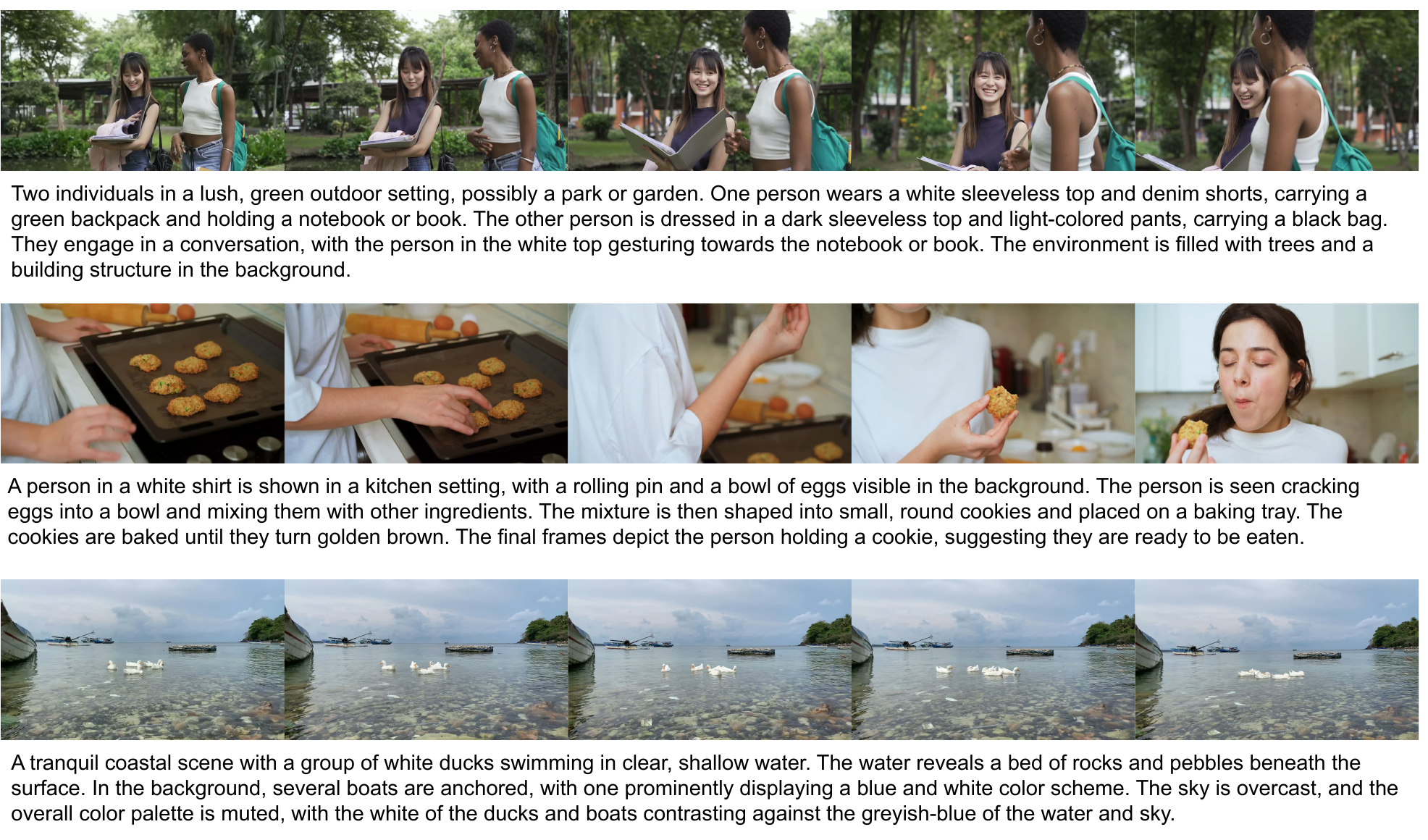}
    \caption{Additional example text captions generated using our xGen-MM-based video captioner. We use such generated video-text pairs for the video generation training.}
    \label{fig:more_caption_examples}
\end{figure}

\clearpage

\begin{figure*}[t]
  \centering
    \includegraphics[width=0.99\linewidth]{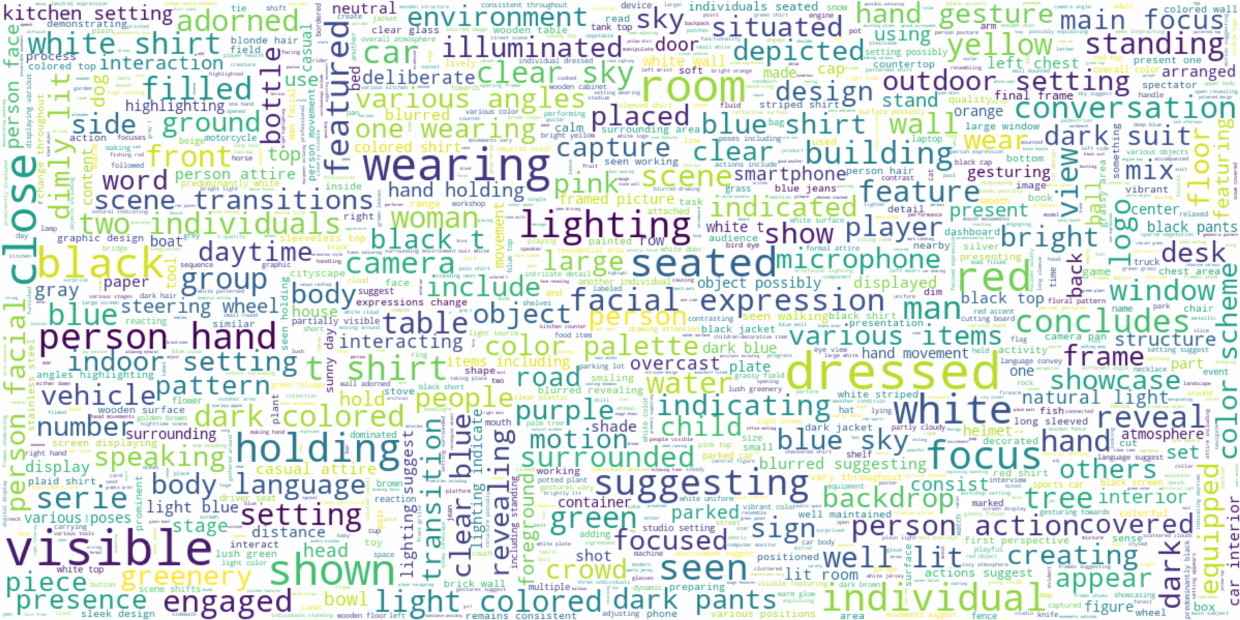}
    \caption{Word cloud of caption samples}\label{fig:caption_word_cloud}
   \end{figure*}

\clearpage

\begin{figure}[tb]
\centering
    \includegraphics[width=\textwidth]{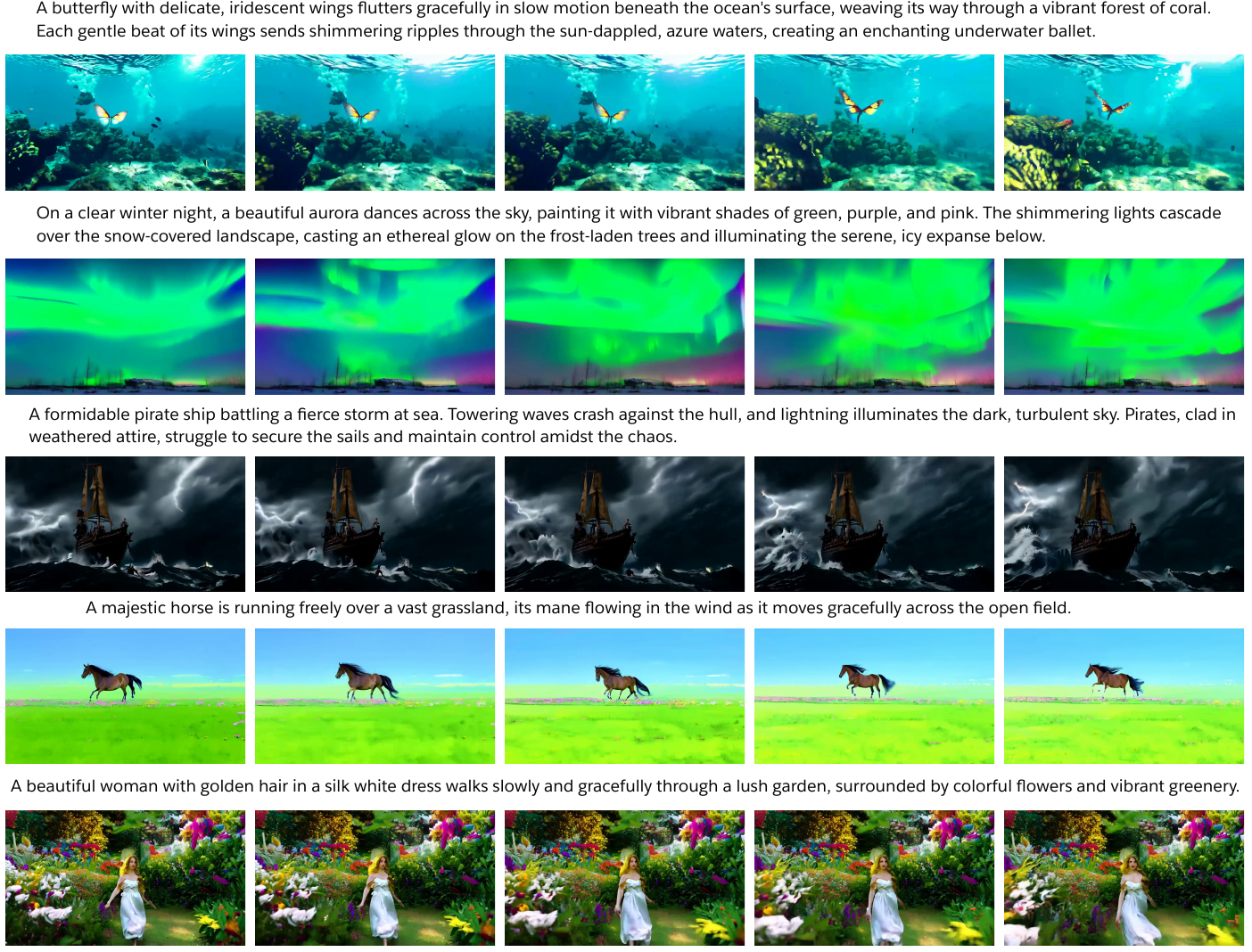}
    \caption{Additional example 720p text-to-video generation results by our xGen-VideoSyn-1 model}\label{fig:result-2-sup}
\end{figure}

\end{document}